\documentclass[11pt,a4paper]{article}
\usepackage[dvipdfmx]{graphicx}
\usepackage[hyperref]{acl2019}
\usepackage{times}
\usepackage{latexsym}
\usepackage{breakurl}
\usepackage{amsfonts}
\usepackage{amsmath}
\usepackage{mathtools}
\usepackage{bm}
\usepackage{lscape}
\usepackage{algorithm,algpseudocode}
\usepackage{xcolor}
\usepackage{ulem}
\usepackage{caption}
\usepackage{setspace}
\usepackage{CJKutf8}
\usepackage[utf8]{inputenc}
\usepackage[T1]{fontenc}
\newcommand{\ja}[1]{\begin{CJK}{UTF8}{ipxm}#1\end{CJK}}

\usepackage[hires]{bxcoloremoji} %
\usepackage{booktabs}

\aclfinalcopy %
\def\aclpaperid{22} %

\title{NTT's Machine Translation Systems for WMT19 Robustness Task}

\author{Soichiro Murakami$^{1}$\thanks{\ \ Equal contribution.}\ , Makoto Morishita$^{2*}$, Tsutomu Hirao$^{2}$ \and Masaaki Nagata$^{2}$ \\
  $^{1}$ Service Innovation Department, NTT DOCOMO, INC., Japan \\
  $^{2}$ NTT Communication Science Laboratories, NTT Corporation, Japan \\
  \texttt{souichirou.murakami.cr@nttdocomo.com}\\
  \texttt{\{makoto.morishita.gr, tsutomu.hirao.kp, }\\
  \texttt{masaaki.nagata.et\}@hco.ntt.co.jp}
  }

\date{}

\begin{document}
\maketitle
\begin{abstract}
This paper describes NTT's submission to the WMT19 robustness task.
This task mainly focuses on translating noisy text (e.g., posts on Twitter), which presents different difficulties from typical translation tasks such as news.
Our submission combined techniques including utilization of a synthetic corpus, domain adaptation, and a placeholder mechanism, which significantly improved over the previous baseline.
Experimental results revealed the placeholder mechanism, which temporarily replaces the non-standard tokens including emojis and emoticons with special placeholder tokens during translation, improves translation accuracy even with noisy texts.
\end{abstract}

\section{Introduction}
\label{sec:intro}
This paper describes NTT's submission to the WMT 2019 robustness task \cite{wmt19robustness}.
This year, we participated in English-to-Japanese (En-Ja) and Japanese-to-English (Ja-En) translation tasks with a constrained setting, i.e., we used only the parallel and monolingual corpora provided by the organizers.

The task focuses on the robustness of Machine Translation (MT) to noisy text that can be found on social media (e.g., Reddit, Twitter).
The task is more challenging than a typical machine translation task like the news translation tasks \cite{bojar18wmt} due to the characteristics of noisy text and the lack of a publicly available parallel corpus \cite{michel18mtnt}.
Table~\ref{tab:example} shows example comments from Reddit, a discussion website.
Text on social media usually contains various noise such as (1) abbreviations, (2) grammatical errors, (3) misspellings, (4) emojis, and (5) emoticons.
In addition, most provided parallel corpora are not related to our target domain, and the amount of in-domain parallel corpus is still limited as compared with parallel corpora used in the typical MT tasks \cite{bojar18wmt}.

To tackle this {\it non-standard} text translation with a low-resource setting, we mainly use the following techniques.
First, we incorporated a placeholder mechanism \cite{crego2016systrans} to correctly copy special tokens such as emojis and emoticons that frequently appears in social media.
Second, to cope with the problem of the low-resource corpus and to effectively use the monolingual corpus, we created a synthetic corpus from a target-side monolingual corpus with a target-to-source translation model.
Lastly, we fine-tuned our translation model with the synthetic and in-domain parallel corpora for domain adaptation.

The paper is organized as follows.
In Section~\ref{sec:system_details}, we present a detailed overview of our systems.
Section~\ref{sec:experiments} shows experimental settings and main results,
and Section~\ref{sec:analysis} provides an analysis of our systems.
Finally, Section~\ref{sec:conclusion} draws a brief conclusion of our work for the WMT19 robustness task.

\begin{table}[t]
\centering
\begin{tabular}{@{}l@{ }l@{}} \\ 
\toprule
(1) & I'll let you know \underline{bro}, \underline{thx}\\
(2) & She had \underline{a ton of} rings.\\
(3) & oh my god  it’s \underline{beatiful} \\
(4) & Thank you so much for all your advice!!\underline{\coloremojiucs{1F62D}\coloremojiucs{1F495}}\\
(5) & \underline{$(\backslash\ast\acute{}\ \ \forall\ \ \grave{} \ast\ )$} so cute \\ %
\bottomrule
\end{tabular}
\caption{Example of comments from Reddit.}
\label{tab:example}
\end{table}

\section{System Details}
\label{sec:system_details}
In this section, we describe the overview and features of our systems:
\begin{itemize}
  \item Data preprocessing techniques for the provided parallel corpora (Section \ref{sec:preprocess}).
  \item Synthetic corpus, back-translated from the provided monolingual corpus, and noisy data filtering for its data. (Section \ref{sec:monolingual}).
  \item Placeholder mechanism to handle tokens that should be copied from a source-side sentence (Section \ref{sec:placeholder}).
\end{itemize}

\subsection{NMT Model}
\label{sec:nmt_model}
Neural Machine Translation (NMT) has been making remarkable progress in the field of MT \cite{bahdanau15alignandtranslate,luong15emnlp}.
However, most existing MT systems still struggle with noisy text and easily make mistranslations \cite{Belinkov:18},
though the Transformer has achieved the state-of-the-art performance in several MT tasks \cite{vaswani17transformer}.

In our submission system, we use the Transformer model \cite{vaswani17transformer} without changing the neural network architecture as our base model to explore strategies to tackle the robustness problem.
Specifically, we investigate how its noise-robustness against the noisy text can be boosted by introducing preprocessing techniques and a monolingual corpus in the experiments.

\subsection{Data Preprocessing}
\label{sec:preprocess}
For an in-domain corpus, the organizers provided the MTNT (Machine Translation of Noisy Text) parallel corpus \cite{michel18mtnt}, which is a collection of Reddit discussions and their manual translations.
They also provided relatively large out-of-domain parallel corpora, namely KFTT (Kyoto Free Translation Task) \cite{neubig11kftt}, JESC (Japanese-English Subtitle Corpus) \cite{pryzant17jesc}, and TED talks \cite{cettolo12wit3}.
Table~\ref{tab:corpus_stats} shows the number of sentences and words on the English side contained in the provided parallel corpora.

\begin{table}[tb]
\centering
\begin{tabular}{lrr}
\toprule
     & \textbf{\# sentences} & \textbf{\# words}\\ \midrule
MTNT (for En-Ja) & 5,775 & 280,543\\
MTNT (for Ja-En) & 6,506 & 128,103\\ \midrule
KFTT & 440,288 & 9,737,715\\
JESC & 3,237,376 & 21,373,763\\
TED  & 223,108 & 3,877,868\\
\bottomrule
\end{tabular}
\caption{The number of training sentences and words on the English side contained in the provided parallel corpora.}
\label{tab:corpus_stats}
\end{table}

\newcite{yamamoto16ialp} pointed out that the KFTT corpus contains some inconsistent translations.
For example, Japanese era names are only contained in the Japanese side and not translated into English.
We fixed these errors by the script provided by \newcite{yamamoto16ialp}\footnote{\url{https://github.com/kanjirz50/mt_ialp2016/blob/master/script/ja_prepro.pl}}.

We use different preprocessing steps for each translation direction.
This is because we need to submit tokenized output for En-Ja translation, thus it seems to be better to tokenize the Japanese side in the same way as the submission in the preprocessing steps, whereas we use a relatively simple method for Ja-En direction.

For Ja-En, we tokenized the raw text into subwords by simply applying {\tt sentencepiece} with the vocabulary size of 32,000 for each language side \cite{kudo18acl,kudo18sentencepiece}.
For En-Ja, we tokenized the text by KyTea \cite{neubig11aclshort} and the Moses tokenizer \cite{koehn07moses} for Japanese and English, respectively.
We also truecased the English words by the script provided with Moses toolkits\footnote{\url{https://github.com/moses-smt/mosesdecoder/blob/master/scripts/recaser/truecase.perl}}.
Then we further tokenized the words into subwords using joint Byte-Pair-Encoding (BPE) with 16,000 merge operations\footnote{Normally, Japanese and English do not share any words, thus using joint BPE does not seem effective. However, for this dataset, we found that Japanese sentences often include English words (e.g., named entities), so we use joint BPE even for this language pair.} \cite{sennrich16subword}.

\subsection{Monolingual Data}
\label{sec:monolingual}
In addition to both the in-domain and out-of-domain parallel corpora, the organizers provided a MTNT monolingual corpus, which consists of comments from the Reddit discussions.
Table~\ref{tab:mono_corpus_stats} shows the number of sentences and words contained in the provided monolingual corpus.

\begin{table}[tb]
\centering
\begin{tabular}{lrr}
\toprule
     & \textbf{\# sentences} & \textbf{\# words}\\ \midrule
MTNT (Japanese) & 32,042 & 943,208\\
MTNT (English) & 81,631 & 3,992,200\\
\bottomrule
\end{tabular}
\caption{The number of training sentences and words contained in the provided monolingual corpus.}
\label{tab:mono_corpus_stats}
\end{table}

\begin{figure*}[ht]
    \centering
    \includegraphics[scale=0.75, clip]{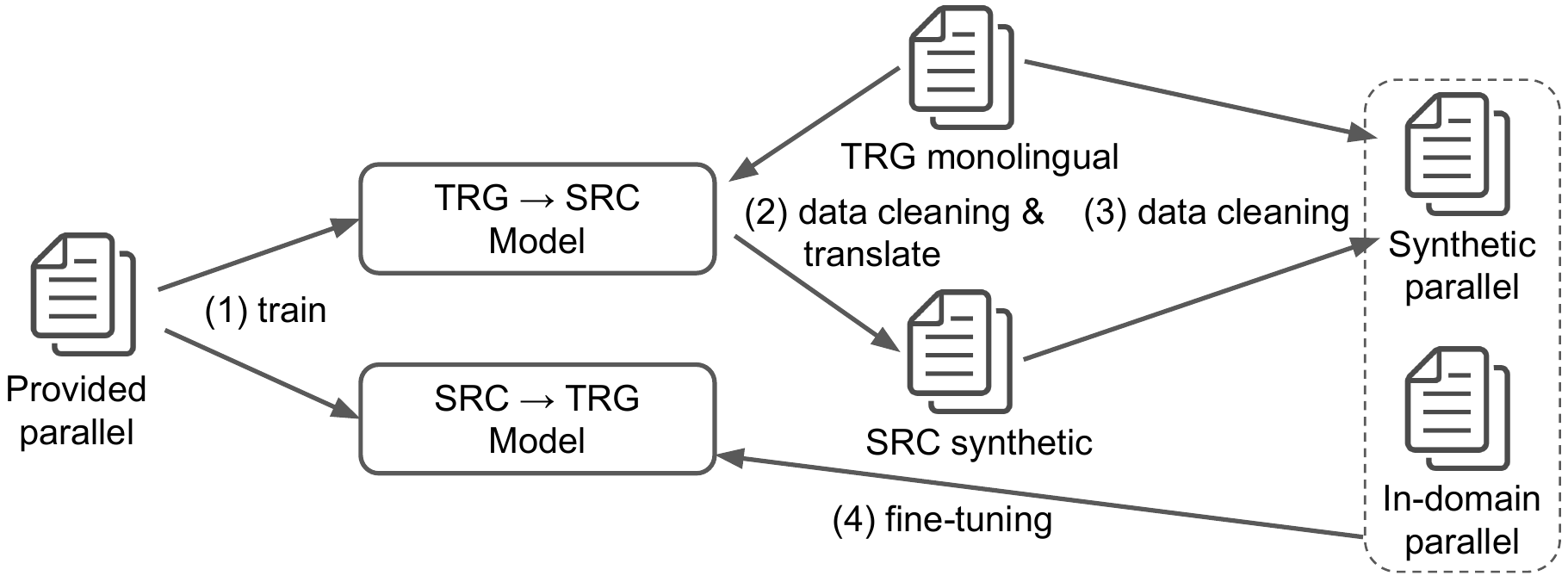}
    \caption{Overview of back-translation and fine-tuning.}
    \label{fig:finetuning_overview}
\end{figure*}

As NMT can be trained with only parallel data, utilizing a monolingual corpus for NMT is a key challenge to improve translation quality for low-resource language pairs and domains.
\newcite{sennrich16acl} showed that training with a synthetic corpus, which is generated by translating a monolingual corpus in the target language into the source language, effectively works as a method to use a monolingual corpus.
Figure~\ref{fig:finetuning_overview} illustrates an overview of the back-translation and fine-tuning processes we performed.
(1) We first constructed both of source-to-target and target-to-source translation models with the provided parallel corpus.
(2) Then, we created a synthetic parallel corpus through back-translation with the target-to-source translation model.
(3) Next, we applied filtering techniques to the synthetic corpus to discard noisy synthetic sentences.
(4) Finally, we fine-tuned the source-to-target model on both the synthetic corpus and in-domain parallel corpus.

Before the back-translation, we performed several data cleaning steps on the monolingual data to remove the sentences including ASCII arts and sentences that are too long or short.
To investigate whether each sentence contains ASCII art or not, we use a word frequency-based method to detect ASCII arts.
Since ASCII arts normally consist of limited types of symbols, the frequency of specific words in a sentence tends to be locally high if the sentence includes an ASCII art.
Therefore, we calculate a standard deviation of word frequencies in each sentence of monolingual data to determine whether a sentence is like ASCII arts.
More specifically, we first define a word frequency list ${\bf x_i}$ of the sentence $i$.
For example, the word frequency list is denoted as ${\bf x}_i = [1, 1, 1, 1, 1]$ for the sentence $i$, ``{\it That 's pretty cool .}''
but as ${\bf x}_j = [1, 1, 1, 1, 3]$ for another sentence $j$, ``{\it THIS IS MY LIFE ! ! !}''.
Note that the length of the list ${\bf x}_i$ is equal to the vocabulary size of the sentence $i$ or $j$ and each element of the list corresponds to the word frequency of a specific word.
Second, we calculate the standard deviation $\sigma_i$ of the word frequency list ${\bf x}_i$ for the sentence $i$.
Finally, if $\sigma_i$ is higher than a specific threshold, we assume that the sentence $i$ contains an ASCII art and discard it from the monolingual data.
We set the threshold to 6.0.

Moreover, since the provided monolingual data includes lines with more than one sentence, we first performed the sentence tokenization using the {\tt spaCy}\footnote{\url{https://spacy.io}} toolkit.
After that, we discarded the sentences that are either longer than 80 tokens or equal to 1 token.

Since a synthetic corpus might contain noisy sentence pairs, previous work shows that an additional filtering technique helps to improve accuracy \cite{morishita18wmt}.
We also apply a filtering technique to the synthetic corpus as illustrated in (3) in Figure~\ref{fig:finetuning_overview}.
For this task, we use the {\tt qe-clean}\footnote{\url{https://github.com/cmu-mtlab/qe-clean}} toolkit, which filtered out the noisy sentences on the basis of a word alignment and language models by estimating how correctly translated and natural the sentences are \cite{denkowski12wmt}.
We train the word alignment and language models by using KFTT, TED, and MTNT corpora\footnote{Note that the JESC corpus is relatively noisy, thus we decided not to use it for cleaning.}.
We use {\tt fast\_align} for word alignment and {\tt KenLM} for language modeling \cite{dyer13fastalign,heafield11kenlm}.

\subsection{Placeholder}
\label{sec:placeholder}
Noisy text on social media often contains tokens
that do not require translation such as emojis,
``\coloremojiucs{1F604}, \coloremojiucs{1F60E}, \coloremojiucs{2665}'',
and emoticons,
``m(\_\ \_)m, %
$(\ \grave{} \cdot \omega \cdot \acute{}\ )$,  %
$\backslash(\ \hat{}\ o\ \hat{}\ )/$''. %
However, to preserve the meaning of the input sentence that contains emojis or
emoticons, such tokens need to be output to the target language side.
Therefore, we simply copy the emojis and emoticons from a source language
to a target language with a placeholder mechanism \cite{crego2016systrans},
which aims at alleviating the rare-word problem in NMT.
Both the source- and target-side sentences containing either emojis or emoticons
need to be processed for the placeholder mechanism.
Specifically, we use a special token ``$<$PH$>$'' as a placeholder
and replace the emojis and emoticons in the sentences with the special tokens.
To leverage the placeholder mechanism, we need to
recognize which tokens are corresponding to emojis or emoticons in advance.
Emojis can easily be detected on the basis of Unicode Emoji Charts\footnote{\url{https://unicode.org/emoji/charts}}.
We detect emoticons included in both the source- and the target-side sentences
with the {\tt nagisa}\footnote{\url{https://github.com/taishi-i/nagisa}} toolkit,
which is a Japanese morphological analyzer that can also be used as an emoticon detector for Japanese and English text.
Moreover, we also replace ``$>$'' tokens at the beginning of the sentence with the placeholders because ``$>$'' is commonly used as a quotation mark in social media posts and emails and does not require translation.

\subsection{Fine-tuning}
\label{sec:fine-tuning}
Since almost all the provided corpora are not related to our target domain, it is natural to adapt the model by fine-tuning with the in-domain corpora.
Whereas we use both the MTNT and synthetic corpora for Ja-En, we only use the MTNT corpus for En-Ja because the preliminary experiment shows that synthetic corpus does not help to improve accuracy for the En-Ja direction.
We suspect this is due to the synthetic corpus not having sufficient quality to improve the model.

\section{Experiments}
\label{sec:experiments}

\subsection{Experimental Settings}
\label{sec:experimental_settings}
We used the Transformer model with six blocks.
Our model hyper-parameters are based on {\it transformer\_base} settings, where the word embedding dimensions, hidden state dimensions, feed-forward dimensions and number of heads are 512, 512, 2048, and 8, respectively.
The model shares the parameter of the encoder/decoder word embedding layers and the decoder output layer by three-way-weight-tying \cite{press17tying}.
Each layer is connected with a dropout probability of 0.3 \cite{srivastava14dropout}.
For an optimizer, we used Adam \cite{kingma14adam} with a learning rate of 0.001, $\beta_{1}=0.9$, $\beta_{2}=0.98$.
We use a root-square decay learning rate schedule with a linear warmup of 4000 steps \cite{vaswani17transformer}.
We applied mixed precision training that makes use of GPUs more efficiently for faster training \cite{micikevicius18mixed}.
Each mini-batch contains about 8000 tokens (subwords), and we accumulated the gradients of 128 mini-batches for an update \cite{ott18scaling}.
We trained the model for 20,000 iterations, saved the model parameters each 200 iterations, and took an average of the last eight models\footnote{The number of iterations might seem to be too low. However, \newcite{ott18scaling} showed that we could train the model with a small number of iterations if we use a large mini-batching. We also confirmed the model had already converged with this number of iterations.}.
Training took about 1.5 days to converge with four NVIDIA V100 GPUs.
We compute case-sensitive BLEU scores \cite{papineni02bleu} for evaluating translation quality\footnote{We report the scores calculated automatically on the organizer's website \url{http://matrix.statmt.org/}.}.
All our implementations are based on the {\tt fairseq}\footnote{\url{https://github.com/pytorch/fairseq}} toolkit \cite{ott19fairseq}.

After training the model with the whole provided parallel corpora, we fine-tuned it with in-domain data.
During fine-tuning, we used almost the same settings as the initial training setup except we changed the model save interval to every three iterations and continued the learning rate decay schedule.
For fine-tuning, we trained the model for 50 iterations, which took less than 10 minutes with four GPUs.

When decoding, we used a beam search with the size of six and a length normalization technique with $\alpha=2.0$ and $\beta=0.0$ \cite{wu16gnmt}.
For the submission, we used an ensemble of three (En-Ja) or four (Ja-En) independently trained models\footnote{Originally, we planned to submit an ensemble of four for both directions. However, we could train only three models for En-Ja in time. In this paper, we also report the score of ensembles of four for reference.}.

\subsection{Experimental Results}
\label{sec:experimental_results}

\begin{table*}[tb]
\centering
\begin{tabular}{lrlrl}
\toprule
                  & \textbf{Ja-En} &  & \textbf{En-Ja} &  \\ \midrule
Baseline model     & 10.8 & & 14.3 &  \\
+ placeholders      & 12.2 & (+1.4) & 15.0 & (+0.7) \\
\ \ \ + fine-tuning      & 11.9 & (+1.1)  & 16.2 & (+1.9) \\
\ \ \ \ \ \ + synthetic      & 14.0 & (+3.2) & --- &  \\
\ \ \ \ \ \ \ \ \ + 4-model ensemble & 14.9 & (+4.1) & 17.0 & (+2.7) \\ \midrule
Submission & 14.8  & & 17.0 &  \\
\bottomrule
\end{tabular}
\caption{Case-sensitive BLEU scores of provided blind test sets. The numbers in the brackets show the improvements from the baseline model.}
\label{tab:bleu}
\end{table*}

Table~\ref{tab:bleu} shows the case-sensitive BLEU scores of provided blind test sets.
Replacing the emoticons and emojis with the placeholders achieves a small gain over the baseline model, which was trained with the provided raw corpora.
Also, additional fine-tuning with in-domain and synthetic corpora also leads to a substantial gain for both directions.
For Ja-En, although we failed to improve the accuracy by fine-tuning the MTNT corpus only, we found that the fine-tuning on both the in-domain and synthetic corpora achieves a substantial gain.
We suspect this is due to overfitting, and modifying the number of iterations might alleviate this problem.
As described in Section~\ref{sec:fine-tuning}, we did not use the synthetic corpus for the En-Ja direction.
For the submission, we decoded using an ensemble of independently trained models, which boosts the scores.

\section{Analysis}
\label{sec:analysis}

\begin{table}[tb]
\centering
\begin{tabular}{lrrr}
\toprule
      & \textbf{Improved} & \textbf{Degraded} & \textbf{Unchanged} \\ \midrule
Ja-En & 9 (53\%) & 0 (0\%) & 8 (47\%) \\
En-Ja & 14 (82\%) & 1 (6\%) & 2 (12\%) \\
\bottomrule
\end{tabular}
\caption{The number of improved/degraded sentences by applying the placeholder mechanism compared with the baseline model.
We manually evaluated all sentences containing placeholders in terms of whether the emojis and emoticons are correctly copied to the output.}
\label{tab:human_eval_placeholder}
\end{table}

\begin{table*}[tb]
\centering
\scalebox{0.8}{
\begin{tabular}{ll}
\toprule
Input          & Woah woah, hang on a minute, let’s hear this guy out.  Amazing title \coloremojiucs{1F602} \\ \midrule
Reference      & \ja{おいおい、ちょっと待てよ。こいつの言うことを聞いてみようぜ。凄いタイトルだ\coloremojiucs{1F602}} \\ \midrule
Baseline       & \ja{うわぁ ちょっと 待 っ て こいつ の 話 を 聞 い て み ま しょ う 驚 く よう な 名前 だっ た わ ね} \\ 
               & (Well wait a minute let's listen to this story It was an amazing name) \\ \midrule
+ placeholders & \ja{ちょっと 待 っ て くださ い この 人 の 話 を 聞 い て み ま しょ う 素晴らし い タイトル だ\coloremojiucs{1F602}} \\ 
               & (Wait a minute, let's hear the story of this person It's a great title \coloremojiucs{1F602}.) \\ \midrule
+ fine-tuning  & \ja{うわー 、 うわー 、 ちょっと 待 っ て 、 この 男 の 話 を 聞 こ う ぜ 。 すご い タイトル だ\coloremojiucs{1F602}} \\
               & (Wow, wow, wait a minute and hear this guy talk. It's an amazing title \coloremojiucs{1F602}.) \\
\bottomrule
\end{tabular}
}
\caption{Translation results on the English-to-Japanese development set. English sentences corresponding to the Japanese translations are also given.}
\label{tab:example_translation_enja}
\end{table*}

\begin{table*}[tb]
\centering
\scalebox{0.8}{
\begin{tabular}{ll}
\toprule
Input          & \ja{>男同士で物言えない奴のただの逆恨み} \\ \midrule
Reference      & >Just misguided resentment from some fellow who can't speak amongst other men.  \\ \midrule
Baseline       & A mere grudge against a man who can't say anything. \\ \midrule
+ placeholders &  > It's just a grudge against guys who can't say anything between men. \\ \midrule
+ fine-tuning  & >it's just inverted resentment for guys who can't say anything between men.\\
\bottomrule
\end{tabular}
}
\caption{Translation results on the Japanese-to-English test set.}
\label{tab:example_translation_jaen}
\end{table*}

\subsection{Effect of Placeholders}
To investigate the effectiveness of using the placeholder mechanism, we compared the translation of the baseline to the model trained with the placeholders.
We manually evaluated how correctly the emojis and emoticons were copied to the output.
Table~\ref{tab:human_eval_placeholder} shows the numbers of sentences on the MTNT test set that are improved/degraded by applying the placeholder mechanism.
These result demonstrate that the placeholder mechanism could improve the translation of the noisy text, which frequently includes emojis and emoticons, almost without degradation.

Tables~\ref{tab:example_translation_enja} and \ref{tab:example_translation_jaen} show examples of translations in the Ja-En and En-Ja tasks, respectively.
Both the emoji (\coloremojiucs{1F602}) and the ``$>$'' token, which represents a quotation mark, were properly copied from the source text to the translation of {\it +placeholders}, whereas the baseline model did not output such tokens as shown in Tables \ref{tab:example_translation_enja} and \ref{tab:example_translation_jaen}.
Thus, we can consider this to be the reason the placeholders contribute to improving case-sensitive BLEU scores over the baseline.

In our preliminary experiments, although we tried a method to introduce the placeholder technique to our systems at the fine-tuning phase, we found that it does not work properly with only the fine-tuning.
This means that an NMT needs to be trained with the corpus pre-processed for the placeholder mechanism before the fine-tuning.

\subsection{Effect of Fine-tuning}
According to the comparison between {\it +fine-tuning} and  {\it baseline} shown in Table \ref{tab:bleu}, fine-tuning on the in-domain and synthetic corpus achieved a substantial gain in both directions.
Accordingly, we can see that the sentence translated by {\it +fine-tuning} has a more informal style than those translated by {\it baseline} and {\it +placeholders} as presented in Tables \ref{tab:example_translation_enja} and \ref{tab:example_translation_jaen}.

\subsection{Difficulties in Translating Social Media Texts}
Challenges still remain to improving the model's robustness against social media texts such as Reddit comments.
As we pointed out in Section~\ref{sec:intro}, various abbreviations are often used.
For example, the term, ``\ja{東スポWeb}'' (literally {\it East Spo Web}) in the MTNT dataset should be translated to ``{\it Tokyo Sports Website}'' according to its reference, but our model incorrectly translated it to ``{\it East Spoweb}''.
Such abbreviations that cannot be translated correctly without prior knowledge, such as ``\ja{東スポWeb} stands for \ja{東京スポーツWebサイト} (literally {\it Tokyo Sports Website})'', are commonly used on social media.

\subsection{Use of Contextual Information}
Some sentences need contextual information for them to be precisely translated.
The MTNT corpus provides comment IDs as the contextual information to group sentences from the same original comment.
We did not use the contextual information in our systems, but we consider that it would help to improve translation quality as in previous work \cite{tiedemann-scherrer-2017-neural, bawden-etal-2018-evaluating}.
For example, in the following two sentences, ``{\it Airborne school isn't a hard school.}'' and ``{\it Get in there with some confidence!}'', which can be found in the MTNT corpus and have the same comment ID, we consider that leveraging their contextual information would help to clarify what ``{\it there}'' means in the latter and to translate it more accurately.

\section{Conclusion}
\label{sec:conclusion}
In this paper, we presented NTT's submission to the WMT 2019 robustness task.
We participated in the Ja-En and En-Ja translation tasks with constrained settings.
Through experiments, we showed that we can improve translation accuracy by introducing the placeholder mechanism, performing fine-tuning on both in-domain and synthetic corpora, and using ensemble models of Transformers.
Moreover, our analysis indicated that the placeholder mechanism contributes to improving translation quality.

In future work, we will explore ways to use monolingual data more effectively, introduce contextual information, and deal with a variety of noisy tokens such as abbreviations, ASCII-arts, and grammar errors.

\section*{Acknowledgments}
We thank two anonymous reviewers for their careful reading and insightful comments and suggestions.

\bibliography{myplain,main}
\bibliographystyle{acl_natbib}

\end{document}